# Robust Real-time Extraction of Fiducial Facial Feature Points using Haar-like Features

## Harry Commin

# Abstract

In this paper, we explore methods of robustly extracting fiducial facial feature points – an important process for numerous facial image processing tasks. We consider various methods to first detect face, then facial features and finally salient facial feature points. Colour-based models are analysed and their overall unsuitability for this task is summarised. The bulk of the report is then dedicated to proposing a learning-based method centred on the Viola-Jones algorithm. The specific difficulties and considerations relating to feature point detection are laid out in this context and a novel approach is established to address these issues.

On a sequence of clear and unobstructed face images, our proposed system achieves average detection rates of over 90%. Then, using a more varied sample dataset, we identify some possible areas for future development of our system.

# Acknowledgements

I would like to thank my supervisor, Prof. Ge, for providing a lab that was available day and night. The enthusiasm and keen work ethic of my lab colleagues was also highly motivating.

For inspiring my now obsessive interest in human visual perception, I thank Dr. Yen Shih-Cheng and his quite exceptional Biological Perception lectures. His insights opened up a massive new perspective on my project and the world I perceive in general.

Finally, thanks go to Santosh Joshi for enriching my year with a colourful mix of academic conversation and extracurricular mischief. The year could not have been the same without the many late-night waffles and general laid-back good humour during times of academic adversity.

# Contents

# Chapter 1

# Introduction

## 1.1 Project Motivation

Extracting salient facial feature points accurately and robustly is absolutely fundamental to the success of the facial image processing system that is built upon them. Such applications include face identification, facial expression recognition, face tracking and lip reading [2]. These feature points generally provide the foundations for the entire system and it is imperative, therefore, that stability and reliability are maximised.

There is no single way to formally define the set of facial feature points, since different applications will have different requirements. Various publications make use of, for example 16 [1], 20 [2] or 34 [3] salient points. In general, these will include distinctive points such as the corners of eyes, mouth and eyebrows – but the specific details vary. Thus, it is desirable to develop an approach that is somewhat adaptable in nature so it can be readily altered to achieve detection of any feature point.

In the past, many systems have ignored this step and relied upon manual marking [3, 4, 5] in order to focus better on the overall system. However, in most cases, it is necessary (or at least highly desirable) to achieve automated detection quickly, or even in real-time. In other words, good detection performance is not the only concern – a system capable of flawless feature point classification may be of little practical use if it cannot meet real-time requirements. So, when choosing our approach, we will try to identify methods that minimise computational complexity wherever possible.

So, the purpose of our work is to develop a method that allows real-time detection of any facial feature point. We seek computational efficiency in order to achieve high frame rate. Where possible, the system should be invariant to general changes within the image, such as scale, translation, rotation and ambient lighting. Additionally, we require robustness regardless of inter-subject differences, such as race, gender and age.

## 1.2 Background

The numerous methods that have been proposed to tackle facial feature point detection problems can be roughly divided into two categories: texture-based and shape-based. Texture-based approaches seek to model the texture properties in the local area around a feature point according to pixel values. Shape-based methods make use of the more general geometries of the face and features, according to some prior knowledge. These two types of approach are by no means independent, and hybrid systems have been shown to give strong results [14, 21].

We will review a few of the more common tools used in approaching the feature point detection problem. This should provide some general insight and aid understanding of the method we finally choose.

### 1.2.1 Colour-based Segmentation

This method is simple and intuitive. Pixels corresponding to skin tend to exist in a relatively tight cluster in colour space, even with variation in race and lighting conditions [11]. A statistical model of skin colour properties is obtained by analysing databases of manually-segmented face images. This model is then used to estimate the likelihood that a given pixel corresponds to skin. Various colour spaces (e.g. RGB, YUV, HSL) and statistical models have been used in an attempt to improve classification performance [8].

Features can then be located within the face through additional local analysis of colour/intensity properties [16], or further image processing techniques such as detection of edges and connectivity [9, 10].

### 1.2.2  Template Matching

In its simplest form, a template can be created by obtaining, for example, an averaged intensity gradient map of a database of frontal face images [17]. Then, detection is achieved by rescaling and shifting this template across a given frame's intensity gradient image, seeking the peak response.

In its basic form, this method is obviously heavily flawed. Searching every possible scale and translation would be inefficient, but omitting any could cause spatial inaccuracies. Furthermore, suitability for feature point detection is limited since there is no particular allowance for variations in expression or pose.

A more sophisticated method is to use deformable templates, which permit constrained inter- and intra-feature deformations according to a number of predefined relationships [7]. However, initial localisation within the image remains an issue and so this method is somewhat better suited to more constrained tasks such as handwritten character recognition [18] and object tracking [7].

### 1.2.3  Feature Tracking

Although feature tracking is very much a separate problem to feature detection, it can provide great support. For example, if the detection system is highly accurate but slow, a high-speed tracking algorithm can speed up overall performance by reducing the number of detection operations. Alternatively, if the detector is somewhat noisy, a tracking technique such as deformable template matching can enforce geometric constraints to improve performance [7]. Other tracking techniques are based on optical flow methods such as the popular Lucas-Kanade algorithm [16].

### 1.2.4  Neural Networks

Neural networks offer exciting potential as feature point detectors as they are very much predisposed to be effective pattern recognisers. This computational model is basically inspired by the way that biological nervous systems, such as the brain, process information. That is, with a large number of highly interconnected, simple elements (neurons) that can function in parallel. These neurons are arranged and connected in a specific way that allows them to solve a single specific task.

For example, a network can be trained to recognise a given pattern (by redistributing input weights and/or altering threshold), such that it will correctly categorise any of the training patterns. However, the real strength of neural networks is how they cope with unfamiliar inputs. By applying a "firing rule" (such as Hamming distance), a neural network can accurately deduce the category that the input pattern most likely belongs to – even if this pattern was not included in training.

An excellent introduction to this topic can be found here: [12]. Further considerations regarding application to feature detection are discussed here: [13].

### 1.2.5  Graph Matching

Unlike many other methods, graph matching offers the possibility to locate a face and the features within it in a single stage. Indeed, the geometrical layout and specific local properties of the features themselves help to directly describe the face.

In general, the attributes of the nodes of the graph are determined by a local image property such as a Gabor wavelet transform [14]. The graph edges describe the spatial displacement between the facial features. The model graph is then matched to the data graph by maximising some graph similarity function. Exhaustively searching for a match is time-consuming, so heuristic algorithms are generally used.

### 1.2.6 Eigenfaces[19]

With eigenfaces, we can begin our discussion of a really crucial topic in pattern recognition – the high-dimensional feature space. This is an important concept, so we will start with the very simple example of a 1x3 pixel greyscale image. Another way of representing this image would be as a single point in 3D space, with the grey level at each pixel determining progression along each axis.

Similarly, a 256x256 greyscale image can be represented as a single point in an abstract 65,536-dimensional space. Now, if we plot a whole database of frontal face images (equally scaled, rotated and centred) in this space, then, due to the similarities between human faces, these points will not just be randomly distributed – they will be somewhat clustered together. The clustering suggests that there is a substantial degree of redundancy in this representation of the face.

This redundancy is important – it means that a relatively small number of vectors are likely to contain a disproportionately large portion of a face's inherent visual characteristics. If we can somehow find a way of selecting these most important vectors, then we can potentially discard all other data and therefore massively reduce the scope of our problem.

In the case of Eigenfaces, the technique for selecting these vectors is Principal Component Analysis (PCA). This method basically treats the training images as a group of 1D vectors, forms their covariance matrix, then finds the eigenvectors (eigenfaces) of this covariance matrix. The subspace described by these vectors therefore provides maximal variance (and therefore minimal reconstruction error of the face images) [20].

Although PCA projections are optimal in terms of correlation, true detection performance is heavily compromised by background conditions, such as lighting and viewing direction. Various improvements and alternative dimensionality-reduction techniques such as Fisher Linear Discriminant (fisherfaces), Indepenent Component Analysis and Support Vector Machines have been proposed [22].

### 1.2.7 Boosting

Boosting is a machine learning algorithm that is often used in training object detection systems. Rather like PCA, the objective of boosting is to select a relatively small number of features that best describe the object we are trying to detect. However, this method uses quite a different approach. With boosting, each feature is analysed individually in terms of its classification performance.

In our context, we have a simple two-class classification problem – for example, "eye" or "non-eye". In order to categorise these, the boosting algorithm learns a number of very simple "weak classifiers" and linearly combines them to form a strong classifier. For our purposes, a weak classifier can be the boolean property of any simple image feature – for example, a thresholded grey level value, or a thresholded filter response centred at a certain pixel.

The great power of boosting is that these weak classifiers alone only need to be slightly correlated with the true classification. For example, taking a database of known "eye" samples and known "non-eye" samples, we may find that using a grey level threshold of 100 at pixel coordinate (3, 5) will give correct classification 51% of the time. Even though this feature is only a little better than random guessing, it can be linearly combined with many other weak classifiers to achieve remarkable detection results.

Obviously, the example above is a very simple one. In practice, our image features should be more sophisticated in order to encode more actual image detail. So, we not only have to consider the boosting algorithm itself, but also the nature of the features we use. These two factors will ultimately determine both detection performance and computational complexity.

### 1.2.8 Discussion of the Human Visual System

While many of the above methods offer very interesting and innovative solutions, it is not necessarily easy to know where to start. So, we will begin by focusing on the fact that we really need to concentrate on elegance and simplicity. To this end, we can take great inspiration from our own visual systems.

The task of facial feature point detection is really rather intuitive to us and yet the range of tools available to our visual systems is somewhat basic. Our brains don't explicitly compute image transforms or edge maps; our monochromatic dark-adapted vision doesn't use any colour information at all. But still we outperform even the best computer vision systems. So, it would be very interesting to briefly consider some of the basic underlying neurological mechanisms it uses.

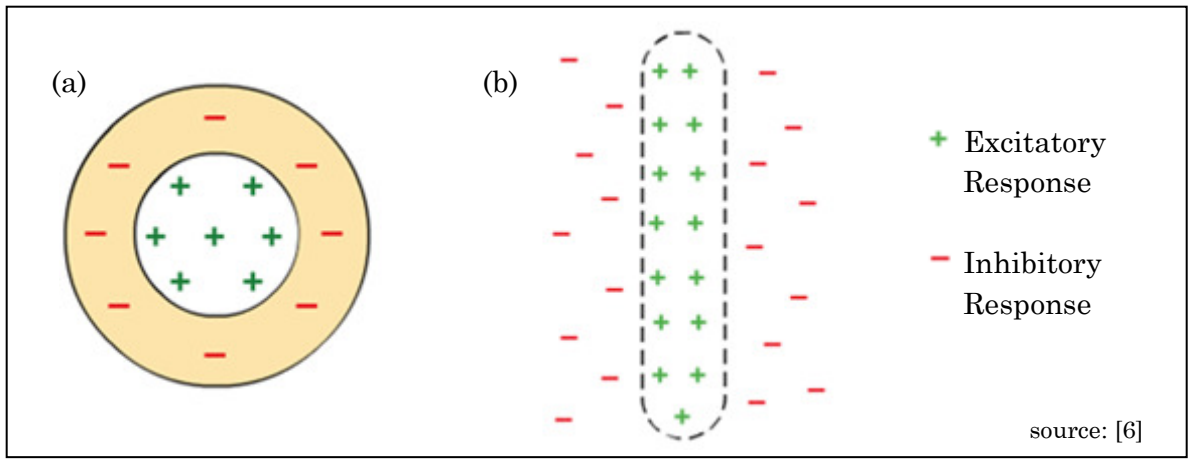

Figure 1.1: (a) *Centre-surround antagonism of a low-level visual neuron.*
(b) *Orientation-sensitive response of a simple cortical cell.*

Neural responses right at the front end of our visual systems (retinal ganglion cells and neurons in the lateral geniculate nucleus), show centre-surround antagonism (see Figure 1.1a). In such cases, an excitatory area of the receptive field is surrounded by an inhibitory area (or vice versa). This means that a small bright circle at the centre of the receptive field will give an excitatory response. Then, as the circle's radius is increased, the response will grow stronger until the inhibitory region becomes stimulated, and response will fall off.

Similarly, in the primary visual cortex (V1), there are simple cortical cells that respond only to stimuli aligned at specific orientations (see Figure 1.1b). These mechanisms are very simple, but are so fundamental to our ability to detect objects that they are often referred to as "feature detectors" in the field of neurophysiology. ([6], pp58-61)

We should be careful to note, however, that our perception of the human face is not simply completed at this early stage of the visual process. Complete facial detection and recognition requires higher-level processes which make sense of those basic early neural responses. In other words, all the information required to solve the detection problem does not have to be explicitly deduced during detection. Instead, some form of prior knowledge can be applied in order to extract meaning (categorise the object).

Using this understanding as a basis, the idea of a learning-based (e.g. boosting) detection scheme suddenly looks particularly exciting. We could choose a family of features that resemble the response of our early visual neurons, and then the learning process would provide us with a fundamental knowledge relating to the feature responses.

# Chapter 2

# Technical Framework

## 2.1 Review of Colour-based Feature Point Detection

### 2.1.1 Research Context (information for Imperial College markers)

The facial expression recognition system of [23] had been developed in the Social Robotics research lab at National University of Singapore. I was provided with the related C++ code project. None of this code was used in the final implementation, but it provided useful learning material. It comprised around 5000 lines, which included large sections unrelated to feature point detection. I was asked to address three main issues:

- The feature point detection system was somewhat unreliable. In particular, it was very sensitive to changes in pose. It was hoped that minor modifications of this method would lead to acceptable performance.
- The code worked in isolation, but needed to be incorporated into the lab's overall robot vision framework in order to be compatible with the other vision components being developed in the lab.
- The code made use of several non-standard libraries. The lab requires that only OpenCV 1.0 and wxWidgets libraries can be used in addition to the standard libraries. OpenCV is an open source computer vision library, originally developed by Intel. Full details can be found here: [38].

This represented some months of work and provided significant insight into the problem at hand, so will be summarised here.

Despite our strong arguments supporting a learning-based feature point detection method, a colour-based approach was not necessarily a bad way to start. This method is intuitive and conceptually simple, and as such provided an insightful introduction to detection, programming with OpenCV and object-oriented programming in general.

The detection algorithm, based on [23], is summarised as follows:

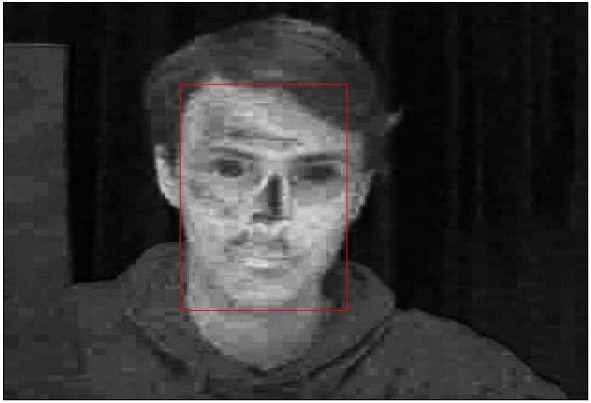

• Figure 2.1 shows the estimated probability that each pixel corresponds to skin, according to the Gaussian Mixed Model described in [23].
• The face region is determined by computing vertical and horizontal histograms of these values.

Figure 2.1: *Locating Face using Skin Likelihood*

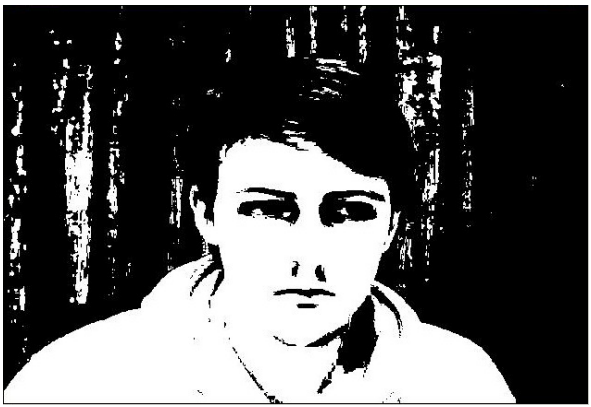

• Having located the face, a thresholded intensity image is used to highlight facial features.

Figure 2.2: *Thresholded Intensity Image*

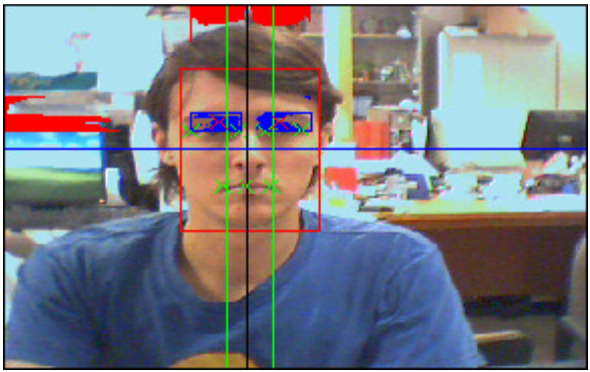

• Histograms (plotted in red) of the below-threshold pixels from the upper part of the face region (marked in blue) are used to define the two eye regions.
• Eye and mouth corners are then marked at the outermost low-intensity pixels in the appropriate regions.
• Eye pupils are marked at the lowest (non-thresholded) intensity value within each eye region.

Figure 2.3: *Feature Point Detection*

Although the performance shown in Figure 2.3 may look reasonable, such results were only achieved under extremely restricted circumstances. The reality is that true performance was unacceptably unstable. The causes for this instability were numerous: pose, lighting conditions and camera exposure were particularly influential.

Extensive effort was put into improving performance using basic methods based on simple pixel properties. Numerous colour models in different colour spaces were tested. Hair-colour models were used in an attempt to separate hair from face.

Morphological 'opening' was used to separate eyebrows from eyes. 'Closing' and connectivity-based segmentation were used in an attempt to classify which feature a pixel belonged to.

In summary, we feel that this task was made particularly difficult by the fact that there was not a single reliable reference. Face height is difficult to define due to variability in fringe hair and exposed neck skin. Determining face width is equally problematic – no single skin-likelihood threshold could consistently separate skin from background objects. This was caused not only by ambient lighting, but also pose relative to the light source. Despite skin being largely homogeneous, the non-planar shape of our faces can give rise to significant variation in the light received by the camera, even within a single frame. It did not seem that any non-adaptive skin colour model could adequately cope with this while, at the same time, non-skin (background) pixels were also regularly misclassified.

The reliability of the face region is fundamental to the success of this approach. Subsequent detection stages rely heavily on facial dimensions in order to define their search regions. Even when these regions are well-defined, this method faces the same problems when refining the search towards feature points.

Intuitively, the problem seemed quite simple, but selecting effective parameters proved to be unbelievably difficult in practice.

### 2.1.2 Relating Back to the Human Visual System

An interesting lesson we can take from this is that our own visual systems can perhaps deceive us into believing that aspects of this detection task are straightforward. Our vision continuously adapts incredibly seamlessly to a wide range of variations. Thus, it becomes difficult for us to obtain an objective and accurate view of what the true nature of the visual stimulus really is. So, perhaps surprisingly, our ability to solve this problem through conscious reasoning seems to be heavily restricted by the immensely powerful subconscious mechanisms that do the job for us.

This really outlines the need for a greater level of abstraction. This is why we need to embed our images in a high-dimensional feature space, and this is why we will finally reject this method.

## 2.2 A Learning-based Approach

We have already discussed several points that make a learning-based detection system a particularly exciting option. For us, there are two approaches that seem to really stand out in addressing all our requirements. These methods are somewhat similar in that they are both adaptations of the ground-breaking Viola-Jones algorithm [25].

So, a strong understanding of the Viola-Jones algorithm is required before we can go on to choosing and fully understanding an eventual method.

### 2.2.1 The Viola-Jones Algorithm

In short, the Viola-Jones method achieves detection at very high speed and has been shown to perform well in detecting various objects, such as faces [25], facial features [32] and pedestrians [26]. There seems to have been little interest in applying this method to feature point detection, but we will see that there is significant potential if the correct restrictions are applied.

There were three really major contributions given by the Viola-Jones method: the "integral image" representation, a modified version of AdaBoost boosting algorithm and a cascaded classifier architecture.

### (i) Integral Image

An integral image is an alternative representation of the standard greyscale image. The point (x,y) in the integral image is simply given by the sum of all points above and to the left of this point in the greyscale image:

$$ii(x, y) = \sum_{x' < x, y' < y} i(x', y')$$

where ii(x,y) is the integral image and i(x',y') is the original image.

The motivation for doing this conversion is that it allows the pixel sum of any rectangular image region to be computed using just four array references (one at each corner). This is immensely efficient and is calculated in constant time regardless of region size.

Viola-Jones then takes advantage of this efficient representation by defining a set of features that are made up of rectangular regions. Some examples of these so-called "Haar-like" features are shown below:

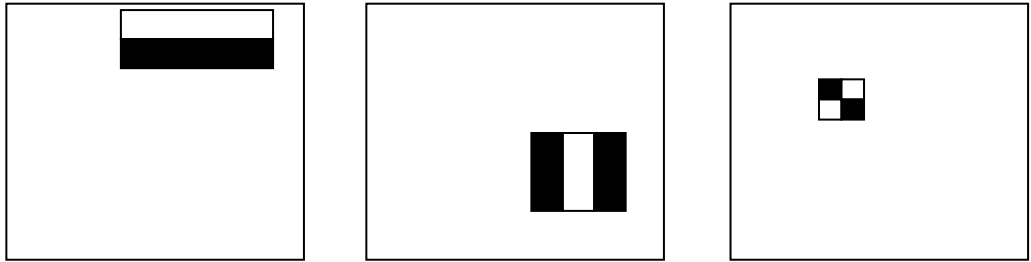

Figure 2.4: *Various Haar-like features within a fixed-size detection window.*

The value of a feature is given by the summed difference between the black and white rectangular regions. It is very important to note that the feature is not just defined by the Haar-like shape alone. Whilst we generally use the word 'feature' to refer to just the Haar-like basis function, strictly speaking, scale and location within the detection window are equally important. So, in order to resolve any potential ambiguity, we will state the full 5-dimensional representation of a feature:

- **Feature Type** – this describes the layout of the chosen Haar-like basis function. The Viola-Jones method uses 3 main categories of feature types: two-, three- and four-rectangle (see Figure 2.4).
- **(x,y)** – exact location (coordinate) corresponding to feature response.
- **(width,height)** – describes the scale of the Haar-like basis function. This should not be confused with the detection window size, which is fixed relative to the Haar-like kernel.

With multiple feature types, scales and translations, the total number of features for a given region will obviously be large. In fact, the number is much larger than the total number of pixels within the detection window. For example, a 24x24 (576-pixel) detection window supports 45,396 Haar-like features. Of course we now need some method of reducing dimensionality in order to select only a small number of features that give the best classification performance.

### (ii) AdaBoost

We introduced the topic of boosting as a method of dimensionality reduction in the 'Background' section. It is now necessary to understand in explicit detail how this algorithm combines results from a simple learning algorithm (weak learner) to form a strong classifier. There is a certain amount of vocabulary we need to familiarise ourselves with, so we will start with a formal definition then work towards a more intuitive understanding.

A formal explanation of the full algorithm [25] is given as follows:

- Given example images $(x_i, y_i), \dots, (x_n, y_n)$ where $y_i = 0, 1$ for negative and positive examples respectively.
- Initialize weights $w_{1,i} = 1/2m, 1/2l$ for $y_i = 0, 1$ respectively, where m and l are the number of negatives and positives respectively.
- For $t = 1, \dots, T$:

  1) Normalise the weights,

  $$w_{t,i} \leftarrow \frac{w_{t,i}}{\sum_{j=1}^n w_{t,j}}$$

  so that $w_t$ is a probability distribution.

  2) For each feature, j, train a classifier $h_j$ which is restricted to using a single feature. The error is evaluated with respect to $w_t$, $\varepsilon_j = \sum_i w_i |h_j(x_i) - y_i|$.

  3) Choose the classifier, $h_t$, with the lowest error $\varepsilon_t$.

  4) Update the weights: $w_{t+1, i} = w_{t, i} \beta_t^{1-e_i}$

  where $e_i = 0$ if example $x_i$ is classified correctly, $e_i = 1$ otherwise and $\beta_t = \frac{\varepsilon_t}{1 - \varepsilon_t}$

- The final strong classifier is:

  $$h(x) = \begin{cases} 1 & \sum_{t=1}^T \alpha_t h_t(x) \geq \frac{1}{2} \sum_{t=1}^T \alpha_t \\ 0 & \text{otherwise} \end{cases}$$

  where $\alpha_t = \log \frac{1}{\beta_t}$

Here, the number of weak classifiers that combine to form a strong classifier is given by T. We will look at how this value is determined when we later discuss the detector cascade.

Another very important point we must highlight is the subtle difference between a feature and a weak classifier (see part (2) of the algorithm). Although we often discuss features and weak classifiers analogously, strictly speaking, a weak classifier is actually what we get as a result of training with a single feature. To clarify, a weak classifier is therefore dependent on feature, threshold and parity:

- *Feature* – the 5-dimensional entity we defined above.
- *threshold* – determined by the weak learner (to minimise misclassifications).
- *parity* – describes whether we subtract the white rectangle pixel sum from the black rectangle pixel sum or vice versa.

Having gained all of the necessary vocabulary and resolved any ambiguities, we should now be able to summarise Viola-Jones' adapted AdaBoost algorithm with total clarity:

---

- Initially, all sample weights are fairly distributed.
i. The weak learner takes a single feature and forms a weak classifier by setting a threshold and parity that best classifies the samples. This is repeated for all features.
ii. The lowest-error weak classifier is selected and applied to the samples.
iii. The weights of the samples are then reallocated to emphasise those that were misclassified.
- Stages (i) – (iii) are repeated T times.
- The strong classifier is formed by a weighted linear combination of the T weak classifiers, with a threshold that yields low error-rate.

---

Since weights are reallocated every time a weak classifier is selected, all weak classifiers must be completely learnt from scratch on each iteration. As we have discussed, the set of features is quite large. The sets of positive and negative samples will also generally number in the thousands. So, even with the great efficiency of the integral image representation, this process requires a very large amount of computation.

Obviously, the number of features to be used for detection is only a relatively tiny subset of those used in training. However, with computational efficiency being so important, Viola-Jones also offers the following cascaded structure to greatly reduce the number of classifiers used during detection.

### (iii) Detection Cascades

The key to the success of the cascade detector structure essentially rests on the following statement:

*It is much easier to say for sure that a given detection region definitely does **not** contain a positive match, than it is to be sure that it definitely **does**.*

So, if we can quickly discard those 'easy' negative matches, more sophisticated detection only needs to be carried out in the more promising image regions.

The cascade does this by performing detection in stages, each one a different strong classifier. The early stages use simple, computationally efficient features to achieve high-speed rejection. The latter stages then use more complex features in order to resolve the more difficult regions and reduce false positive rates.

The idea is that while a positive match will have to pass through every detection stage, this is an extremely rare occurrence. The vast majority of detection regions are rejected at an early stage.

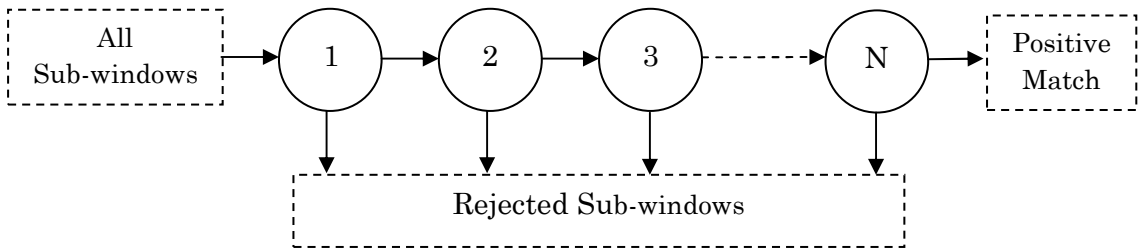

<u>Figure 2.5</u>: *Schematic of an N-stage Detection Cascade.*

Since the false positive rate ('false alarm rate') and successful detection rate ('hit rate') both progress multiplicatively, the classifier thresholds need to be kept low in order to minimise false negatives. What we are now saying is that the objective is no longer to maximise classification with every classifier (as was the case with our original discussion of AdaBoost). The important thing now is that we don't discard any positive matches whilst we quickly reject definite negatives. So the threshold for selecting a strong classifier (initially set at $\sum_{t=1}^{T} \alpha_t$) needs to adapt to accommodate this.

The way this is done is to enforce limits on the minimum hit rate and maximum false alarm rate for each cascade stage during training. For example, we may require a minimum hit rate of 0.995 and a maximum false alarm rate of 0.5. For a 15-stage classifier cascade, the worst-case result would then be an overall false detection rate of $0.5^{15} \approx 3\mathrm{x}10^{-5}$, whilst still maintaining an overall hit rate of $0.995^{15} \approx 0.93$.

In order to accommodate these constraints, the number of weak classifiers per strong classifier, T, has to be allowed to vary. For example, higher maximum hit rates and lower minimum false alarm rates will generally require a larger number of weak classifiers in order to meet these tight constraints. This would, in theory, allow better classification performance per cascade stage, but the trade-off is that this will also increase computation at each stage.

However, we should be mindful of the fact that peak classification performance will ultimately depend on the quality and quantity of our training data. An infinite number of 'perfect' samples does not exist. So, in using looser detection requirements to speed up detection, the actual sacrifice in classification performance may be relatively small.

### 2.2.2 Adaptations of Viola-Jones

Now that we have a strong understanding of the Viola-Jones algorithm, we can finally move on to look briefly at the adaptations of this method. The first one we will consider was proposed by Lienhart et al. [27]. The biggest contribution of this paper was to extend the Haar-like feature set, but it also presented analysis of alternative boosting algorithms.

The major change to the feature set was the inclusion of rotated features. The integral image offered by Viola and Jones only allowed for the fast computation of upright rectangular image areas. So, Lienhart et al. use a second auxiliary image (the Rotated Summed Area Table, RSAT) to achieve fast computation of rectangles offset by an angle of 45° to the horizontal. The details of how the RSAT can be computed in a single pass over the image can be found in [27].

There are also some adapted line features and, interestingly, centre-surround features:

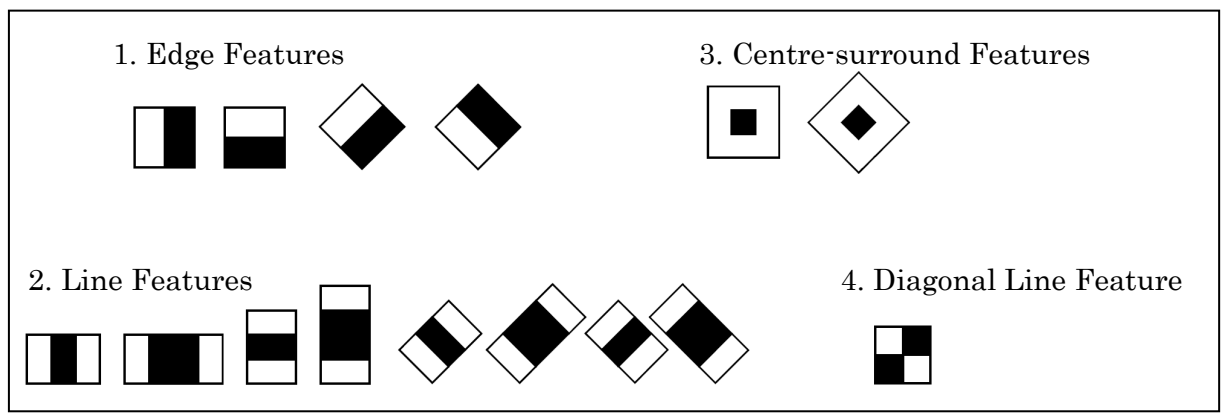

Figure 2.6: *The extended Haar-like feature set proposed in* [27].

Now a fascinating observation is to compare these features with the neural responses we saw earlier in Figure 1.1. The similarities are striking, but in using computationally-efficient rectangular shapes, the Haar-like features are relatively 'blocky' and imprecise in nature.

It is generally accepted that the best way of modelling those neural responses is actually to use Gabor filters [28, 29]. This brings us on to our discussion of the approach adopted by Vukadinovic and Pantic [2]. Their Gabor wavelet-based method was developed specifically for the task of facial feature point detection, with an impressive average 93% detection rate on 20 feature points.

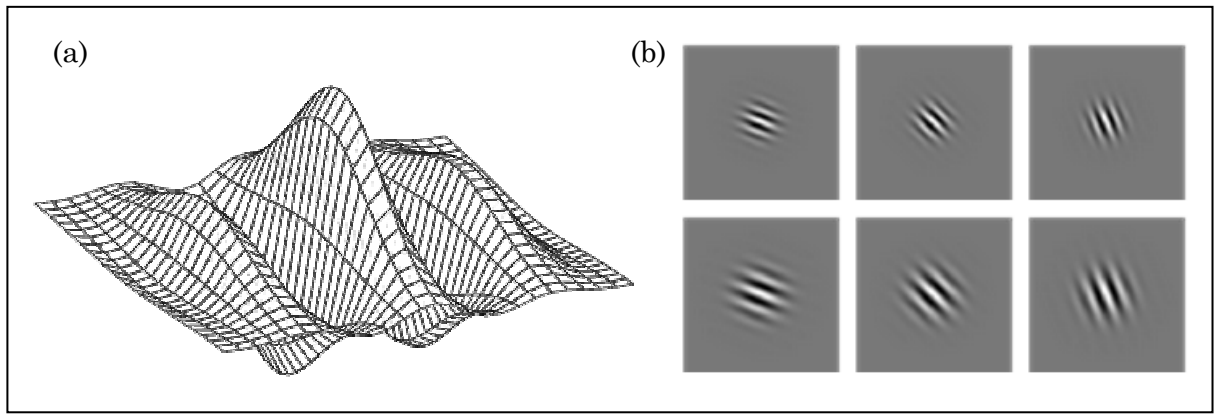

Figure 2.7: (a) *A 3D representation of a Gabor feature.* (<u>Source</u>: [30]).
(b) *Some example Gabor features.* (<u>Source</u>: [31]).

However, there are two key concerns regarding this implementation. Firstly, the non-rectangular nature of the Gabor features means the highly-efficient integral image representation cannot be used. Vukadinovic and Pantic do not make reference to frame rate, but we expect it to be slower than a system based on Haar-like features.

The other concern is the difficulty of practical implementation. Having already spent so long trying to work with colour-based methods, time constraints are extremely tight. To this end, the Haar-like implementation has a major advantage: Lienhart et al. developed their system at Intel and so OpenCV offers support for this method.

So, our implementation will be based on Haar-like features, but we will draw upon some of the methods used by Vukadinovic and Pantic for specific application to feature point detection.

# Chapter 3

# Implementation

Constructing our detector requires two main stages: training then detection. Despite this chronological order, decisions regarding training affect the detector and vice versa. Therefore we will divide our description of the detector so that some aspects can be discussed after we have some understanding of training techniques.

A lot of effort was put into making it quick and easy to add any new detectors to the system. A separate document has been prepared detailing the simple practical steps required to do this – from training to detection (see Appendix A).

## 3.1 Detector Structure

The key to our method rests on the fact that, although facial feature points contain considerably less structural information than a whole face, the background in which we find them is also very restricted. In other words, if we can reliably narrow the search area down to a relatively small region around the feature point, then our task is made much easier.

The way we do this is to perform 3 stages of detection in a hierarchical manner:

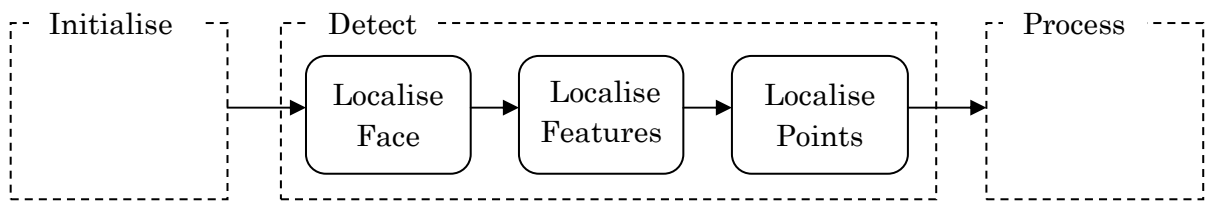

Figure 3.1: *Schematic overview of detector structure.*

In a way, this mimics the cascaded structure of the Viola-Jones algorithm itself. We are saying, for example, that a facial feature definitely will not lie within a non-face, so non-faces should be rejected first. However, the key difference here is that the latter detection stages are not necessarily more complex – they are just more task-specific. The practical implications of this will be discussed in 'Training', below.

## 3.2 Design Considerations

A desirable property of many computer vision systems is rotation-scale-translation (RST) invariance. Our detector is no different – we wish for detection to be achieved regardless of how the face is located, scaled or orientated within the image.

The Viola-Jones method can achieve translation and (most) scale invariance due to the way the whole image is searched at numerous scales. A degree of lighting correction is also achieved through contrast stretching [27]. Rotation invariance is not intrinsically possible, but due to the specific nature of our problem, we can seek to partly address this issue.

In a 3D vision problem, we are concerned with rotation about the three spatial axes:

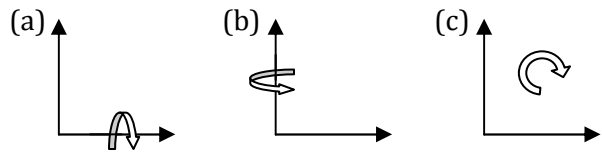

Figure 3.2: *Rotation about (a) x-axis (b) y-axis and (c) z-axis*.

Under normal circumstances, facial rotations about the x-axis don't exceed relatively small nodding motions. Rotation about the y-axis is much more problematic, since the 3D shape of the face is such that this rotation can lead to features being occluded from view. As such, we will leave these for discussion in the "Future Work" section.

However, the problem we can seem to address is rotation about the z-axis (rotations parallel to the image plane). This problem should be relatively easy to solve, since the information available in the image is effectively not changed. All we need to do is detect this rotation, and then we should be able to retrieve facial information in the same way as for an upright face.

We propose to track this rotation about the z-axis by using our detected feature points to provide a reference. In particular, eye corners are distributed in a somewhat horizontal manner across the face and remain relatively fixed under changing facial expression. Thus, by estimating a best-fit line that connects these points, we can compute its angle offset to the horizontal and correct our system accordingly.

Since we are only trying to detect a single line, a basic linear least squares (LLS) approach should be sufficient. We construct the LLS problem by simply rearranging the equation of a straight line into the required form $A\mathbf{x} = b$:

- Eqn. of a straight line: $y = mx + c$

- Rearrange: $y\left(\dfrac{1}{c}\right) + x\left(\dfrac{-m}{c}\right) = 1$

- Form $A\mathbf{x} = b$: $\begin{bmatrix} x_0 & y_0 \\ \vdots & \vdots \\ x_n & y_n \end{bmatrix} \begin{bmatrix} -m/c \\ 1/c \end{bmatrix} = \begin{bmatrix} 1 \\ \vdots \\ 1 \end{bmatrix}$  (where $(x_i, y_i)$ are eye corner points).

The next step is to compute the singular value decomposition (SVD) of the matrix A, to give $A = UDV^T$. The least squares solution for $\mathbf{x}$ is then given by $\mathbf{x} = A^+b$, where $A^+$ is the pseudoinverse of A and is given by:

$$A^+ = V \begin{bmatrix} D_0^{-1} & 0 \\ 0 & 0 \end{bmatrix} U^T$$

(where $D_0$ is the diagonal matrix of the singular values of A).

Having obtained $\mathbf{x} = [\text{-m/c},\ 1/c]^T$, a simple rearrangement yields the gradient of the best fit line, m. Finally, the estimated facial tilt offset angle is given by $\alpha = \tan^{-1}(m)$.

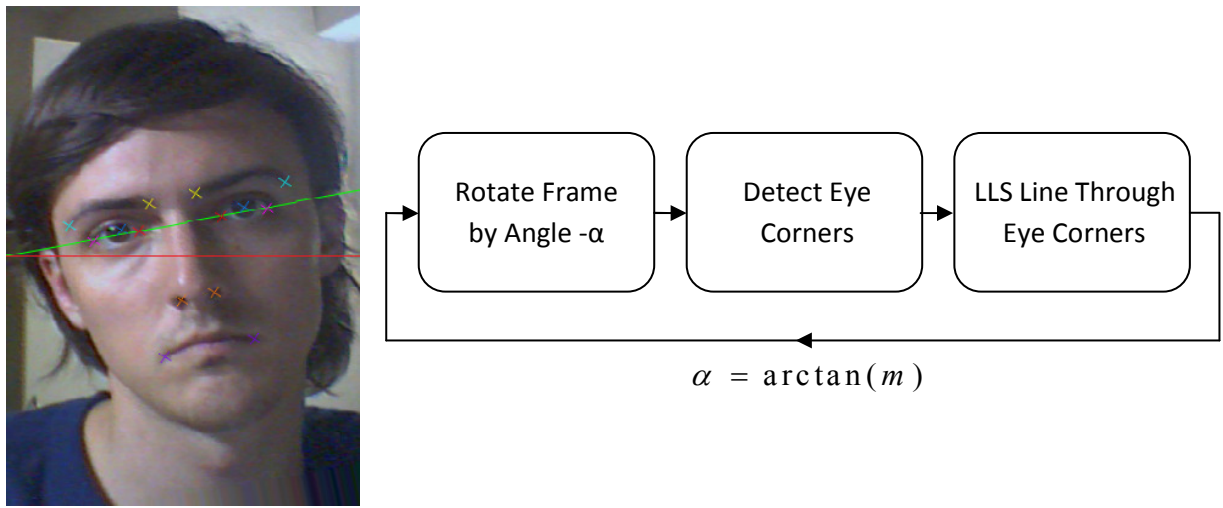

Figure 3.3: *A video capture of tilt detection and a summary of the correction mechanism.*

The tilt-correction mechanism is summarized in Figure 3.3, above. Obviously, a current limitation of this method is that the detection of a relatively upright face must take place before a reference tilt angle can be deduced from the eye corners.

Now, the final consideration we will address is that regarding facial geometry. As mentioned earlier, there is a whole range of shape-based methods that use facial geometry to locate features. Although we have largely left this for discussion in the 'Future Work' section, we did use one small aspect of facial geometry. When only three eye corner points are known, we estimate the fourth through the simple assumption that both eyes are the same width and alignment.

Thus, our full system schematic is as follows:

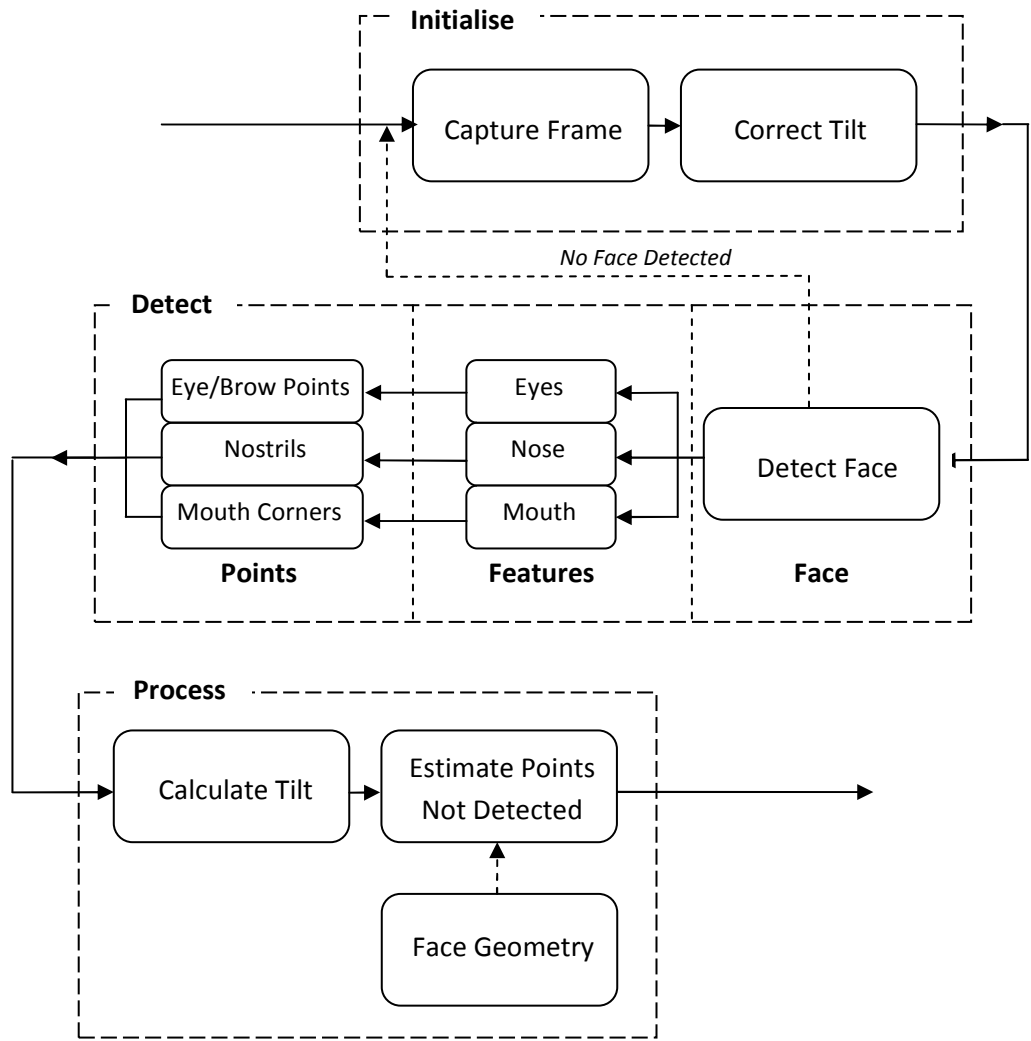

Figure 3.4: *Detection system overview.*

## 3.3 Training

We will now discuss how detection cascades were trained for feature points. In order to concentrate fully on the task of feature point detection, we chose not to train our own face or feature detectors (since samples of these are available in the public domain). The face detector we used was OpenCV's "haarcascade_frontalface_alt2.xml". Detectors for Left Eye, Right Eye, Nose and Mouth, were obtained from [32].

### 3.3.1 Image Database

For the sake of training, we have made extensive use of the BioID Face Database [33] with the FGNet Markup Scheme [34]. The BioID Face Database is a set of 1521 images featuring 23 different subjects. Each image contains a frontal view of a face amongst various indoor surroundings. Race, gender, facial expression, pose and lighting vary. There is further variation in terms of the presence of facial hair and/or spectacles. The faces appear at various locations, scales and (near-upright) rotations within the images.

The FGNet Markup Scheme is an enormous asset. It consists of 1521 data files that describe the precise coordinates of 20 important facial feature points in each of the BioID images. This full database of coordinates was entirely marked by hand, so can be considered to be precise.

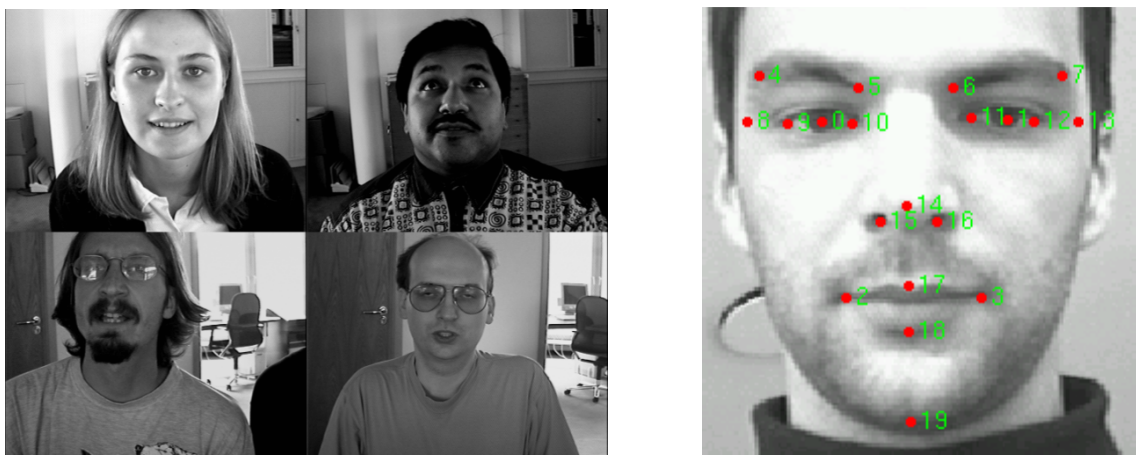

Figure 3.5: (a) *Example BioID Face Images.*          (b) *The FGNet Markup Scheme.*

### 3.3.2 Positive Samples

In our overview of eigenfaces (in 'Background'), we briefly mentioned the importance of equal scale, alignment and rotation of positive training images. Features in the Viola-Jones method are also defined according to their spatial location, so the same principles apply. What we want is for the salient information (which defines the feature point) to be distributed equally in each sample and so Haar-like features stand a better chance of returning similar responses to all positive samples.

During training, the issue of correctly centring the sample is achieved by use of the provided coordinate ground truths. The Viola-Jones method is also intrinsically resistant to variation in ambient light intensity. However, we still have to look at ways to correct tilt and normalise scale. It should be noted that we only do this during the creation of positive samples, since a greater degree of generality is desirable in negative samples. This ensures better rejection of false positive classifications at an early stage.

### (i) <u>Correcting Sample Tilt</u>

We have already discussed the idea of correcting facial tilt during detection. Applying the same process to our database of face images should align the samples as desired.

In our earlier discussion of tilt-correction during detection, the image rotation was achieved using OpenCV functions. However, we now need to rotate our ground truth coordinate values explicitly, so we will discuss a method for doing this [35].

We consider representing a point, P, in x-y and the rotated coordinate space, x'-y':

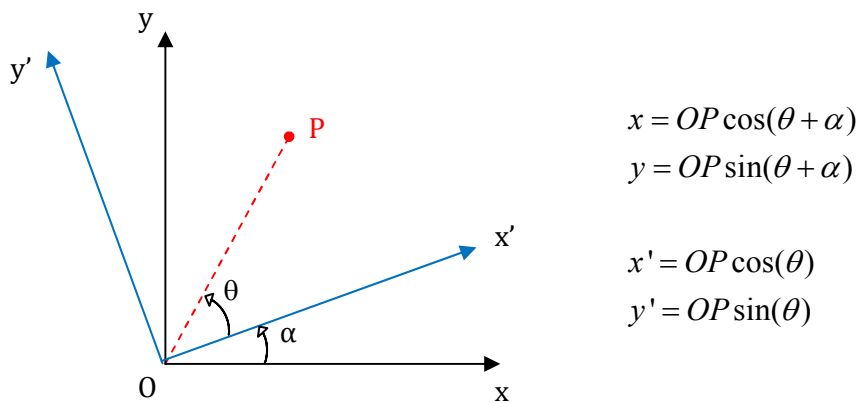

$$x = OP\cos(\theta + \alpha)$$
$$y = OP\sin(\theta + \alpha)$$

$$x' = OP\cos(\theta)$$
$$y' = OP\sin(\theta)$$

<u>Figure 3.6</u>: *Rotating the coordinate axes.*

We can now express the x-y coordinates in terms of the x'-y' coordinates by using the standard trigonometric angle sum relations:

$$\sin(a+b) = \sin(a)\cos(b) + \cos(a)\sin(b)$$
$$\cos(a+b) = \cos(a)\cos(b) - \sin(a)\sin(b)$$

This yields:
$$x = x'\cos\alpha - y'\sin\alpha$$
$$y = x'\sin\alpha + y'\cos\alpha$$

The centre of this rotation is somewhat arbitrary for our purposes. We choose the image centre, since this means the image data that we inevitably discard during a rotation is evenly distributed and minimal. We incorporate this correction term into the above equations to finally yield:

$$x = x_{centre} + (x' - x_{centre})\cos\alpha - (y' - y_{centre})\sin\alpha$$
$$y = y_{centre} + (x' - x_{centre})\sin\alpha + (y' - y_{centre})\cos\alpha$$

A BioID database image, marked with original ground truths is compared with our rotated version below. In each case, the least-squares line through the eye corners is shown.

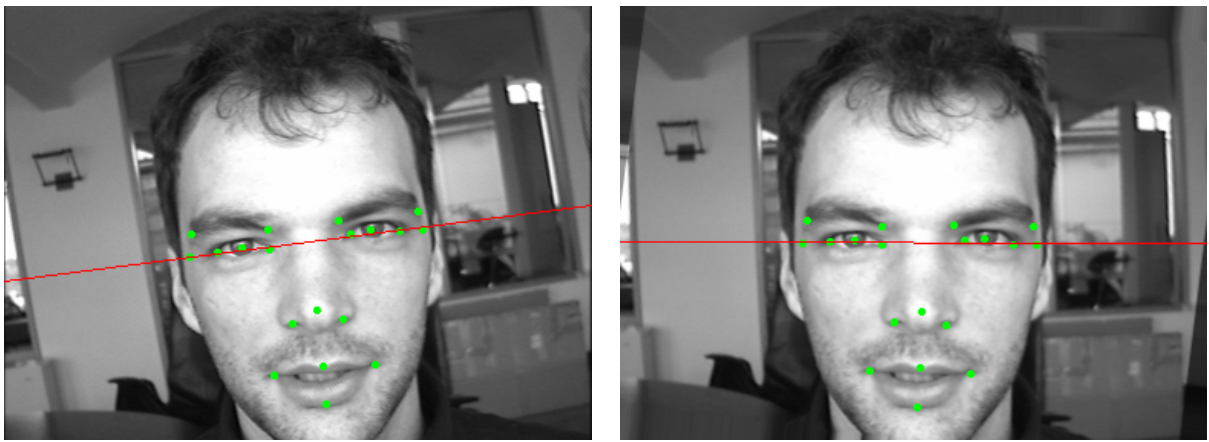

Figure 3.7: *Rotating both image and ground truths.*

We should note at this point that this set of tilt-corrected ground truths contains plenty of information that is useful beyond simply extracting image samples. It is also a large and accurate database that describes facial geometry. Without correcting tilt, this geometrical information is of course accessible, but not in context to our detectors. Although our system adapts to tilt, our detectors essentially operate upright on rotated images. Thus, by applying our face and feature detectors to tilt-corrected images with available ground truths, we were able to look more precisely at where feature points are likely to fall within these regions.

**(ii) Normalising Scale**

In order to normalise sample scales, we need some sort of reference measure that is relatively consistent in all face images. A further consideration is that of local scale; if the face does not lie parallel to the image plane, then features across different depths (within the same face) will appear at inconsistent scales.

For these reasons, we choose our reference to be eye width (magnitude of the distance between inner and outer eye corners). Compared to mouth and brows, eyes stay relatively stationary under varying expression. They are situated at each side of the face, so should give good scale reference on that side.

Next, we have to decide on the approximate amount of facial information we want in each sample. There does not seem to be any definite way to make this decision. Indeed, no publications we found seemed to normalise scale at all. So, we chose to centre our scales around the fixed size proposed by Vukadinovic and Pantic: 13x13.

So, we initially propose to select sample size is as follows:

1. Find the mean magnitude of relevant eye width for all images in database.
2. Rescale these values such that mean = 13.
3. Round to nearest odd integer (to allow a well-defined centre pixel).

This gave us various sample sizes ranging from 7x7 to 33x33. This great variation would seem to illustrate the importance of scale normalisation.

In terms of the information each sample contains, for example, for the inner corner of the left eye, every positive image reached from the extremity of the canthus to the edge of the pupil.

However, in practice, this approach was found to give surprisingly poor results. It seems that applying such a tight restriction on scale makes it difficult to pinpoint the correct scale during detection. The way we addressed this problem was to instead use a small range of image patch sizes for each feature point, then rescale them all to the correct normalised scale. After some trial and error, we decided to have two additional scales – one smaller and one larger (to the nearest odd number):

(a)                                    (b)

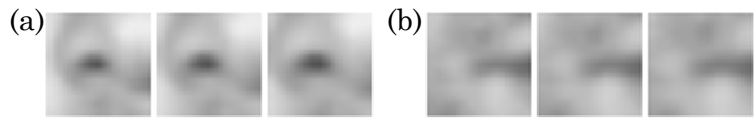

Figure 3.8: *Samples at 3 scales for (a) Nostril and (b) Outer Left Brow.*

This seemed to give the greater generality that we needed, without any significant increase in false positives.

**(iii) Horizontal Mirroring**

The human face displays a degree of horizontal symmetry and this should be exploited. So, instead of training a feature point detector for every facial feature point, we only did so for left-side feature points. Then, during detection, the right-side regions of interest were flipped for detection, and then the detected point could be flipped back.

We should note that flipping image patches during detection may slow detection down. However, the image patches are small and the many days of training computation saved were believed to be more significant. A long-term solution to this problem would be to alter the detection cascade xml files directly. This will be discussed later in 'Future Work'.

Of course, our faces are not perfectly symmetrical, so we could also double the size of our training databases by flipping sample image patches from the right side of the face. However, in practice, we found that this offered no significant increase in performance. This makes sense because the variation of a given feature between the left and right sides of a single face are likely to be tiny compared to the difference between two different people.

One variation that may be offered by left-right sample pairs is that of scale. Any face that is not perfectly frontally-facing will have left and right features at different depths (and therefore different scales). However, we feel that our systematic method of normalising scale in proportion to local feature dimensions and then rescaling explicitly is far more adaptable and rigorous.

**3.3.3 Negative Samples**

As previously mentioned, the background information local to each feature point is very restricted in nature. This is a very powerful statement. In contrast to most standard detection tasks, it is not necessary for us to achieve detection amongst massively varying backgrounds (such as indoor and outdoor scenes). Instead, we just have to deal with the relatively homogenous set of local skin regions.

In order to do this, we have adapted the method used by Vukadinovic and Pantic. In their work, they proposed that 9 positive and 16 negative samples (sized 13x13 pixels) could be taken for each feature point:

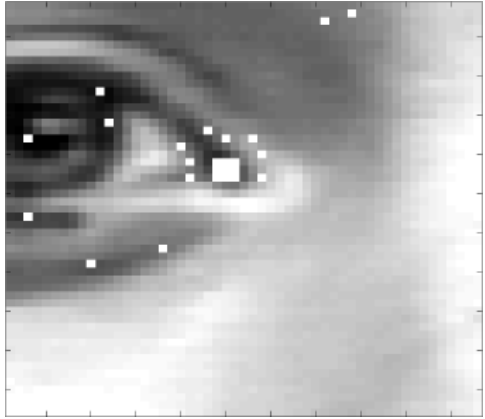

Figure 3.9: *Positive and negative sample selection method, used in* [2].

Their positive samples are in a 3x3 block, centred on the feature point. Then, there are 8 inner and 8 outer negative samples. The inner samples are randomly located at a 2-pixel distance from the positive sample block. The outer samples are randomly distributed in a local region (which we set to be one eye width squared).

The only difference with our method is that we only have one positive sample per image feature. We felt that, after incorporating our method of multiple scaling, we were able to achieve some spatial variation, while using 3 samples rather than 9 (therefore allowing faster training). Note that scale normalisation and tilt correction were only carried out on positive samples, since negative patches need to be rejected at all scales and rotations.

## 3.4 C++ Code Implementation

Please refer to Appendix A for full details on how a detector can be implemented. In summary, any feature or feature point can be passed into the same detection function by simply initialising its following members:

```
CvRect ROI; //general ROI for detection
double HaarParams[4]; //Viola-Jones detection parameters
CvHaarClassifierCascade* cascade; //Haar cascade xml file
bool IsPoint; //for face, eyes etc: want LARGEST region, not point
bool OK; //check if feature found
bool OnRightSide; //flip image patches for right-side features
```

# Chapter 4

# Results

## 4.1 Testing Datasets

In order to carry out thorough quantitative analysis of our system, it would be desirable to perform detection on a marked face database, like the BioID Database. That way, we could make accurate comparisons between ground truths and our estimated feature point locations. However, we cannot use our training database for testing as this will not give us any idea of generalisation.

One could argue that the right-side feature points were not used in training and therefore could be used for testing purposes. However, as we said before, variations between the two sides of a single face are relatively small. Instead, we really need to see if our training database has provided enough generality to perform well on *any* face.

An alternative approach could have been to reserve a subset of the BioID Database just for testing. However, this would have left fewer images available for training and so detection performance may have suffered. Also, with only 23 different subjects in the database, there is not much scope to train and test on different people.

Unfortunately, the only other (free) marked-up database we could find contained just one face. Thus, our testing comprises two main parts:

i)      Quantitative evaluation using a marked-up database of just one face.
ii)     Qualitative analysis using a database containing many different subjects.

The marked-up database we used is the "Talking Face Video" from FGNet [36]. It comprises a sequence of 5000 frames showing a person in conversation. There are 68 points marked on each frame (these include all the points we detected). We used the first 2000 frames.

Our second test database is the Facial Recognition Technology (FERET) colour database [37]. Unfortunately, the FERET database includes many non-frontal images (such as left and right profile views), which are not relevant to our work. So, we first had to segment the database and extract only the frontal images. We used 1175 images from DVD 1 of the FERET Database.

---

Please Note: Due to the large number of images involved, only a few results can be shown in this report. The full results, along with further real-time detection results, can be viewed in video form at: **YouTube.com/HarryComminFYP**

---

## 4.2 Results from Marked-up Database

Since this database is effectively a video sequence, we can incorporate facial tilt correction between frames. We will therefore measure the effectiveness of facial tilt correction by comparing detection scores with and without tilt correction.

One way of classifying a successful detection is by measuring proximity to ground truth value as a proportion of inter-ocular distance (distance between eye centres). In agreement with Vukadinovic and Pantic, we believe that 30% of inter-ocular distance is far too approximate and that 10% is a more meaningful measure.

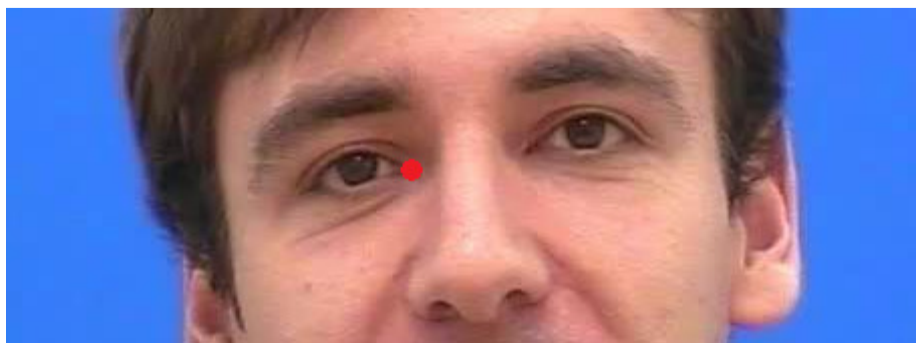

<u>Figure 4.1</u>: *The allowable margin of error for 'success' (10% of inter-ocular distance).*

According to this definition of 'success', we achieved the following results:

| Feature Point | | No Tilt Correction | Full Tilt Correction | Half Tilt Correction |
|---|---|---|---|---|
| Inner | Left Brow | 78.65% | 93.10% | 96.40% |
| Outer | | 99.05% | 95.30% | 97.95% |
| Inner | Right Brow | 99.50% | 96.85% | 99.90% |
| Outer | | 99.30% | 69.65% | 66.90% |
| Outer | Left Eye | 94.35% | 95.20% | 98.20% |
| Pupil | | 92.00% | 92.75% | 94.80% |
| Inner | | 96.05% | 94.45% | 97.45% |
| Inner | Right Eye | 98.15% | 95.30% | 98.20% |
| Pupil | | 92.50% | 92.05% | 94.60% |
| Outer | | 96.35% | 95.00% | 97.35% |
| Left | Nostrils | 98.05% | 94.80% | 97.45% |
| Right | | 98.80% | 96.25% | 98.85% |
| Left | Mouth | 83.00% | 86.55% | 89.60% |
| Right | | 81.85% | 86.50% | 89.85% |
| | | 93.40% | 91.70% | 94.11% |

<u>Figure 4.2</u>: *Success rates for each feature point.*

Overall, results are encouraging. The system largely succeeded above 90% for most feature points. As we would expect, performance regarding mouth corners was below average. This is very much a topic for future development (see Future Work, below).

The performance when using tilt correction was perhaps a little disappointing. It was initially thought that this could be due to unwanted fluctuations in tilt estimation. So, we tried reducing this by only partly correcting tilt (by half). Further investigation is required to see whether these differences are significant. Upon inspection of the videos, it is difficult to see any particular loss of performance when correcting tilt. However, its advantages become more clear for larger tilt, when the face detector begins to fail.

Another thing to point out is the poor detection of the outer right brow. When we inspect the results, there does not seem to be any particular problem with this feature point. Indeed, further analysis reveals that success rate jumps to 98.75% if we allow an error of 15% inter-ocular distance. This detector is just a flipped version of the left brow (which performed well), so we can assume that fluctuations relative to the face are no more significant here. In other words, this detector is stable, but the way it defines a feature point for a given face may differ slightly to the mark-up scheme. For most detection tasks, we feel this would be acceptable. While confirmation on another marked database would be desirable, we don't consider this anomaly to be of special concern.

Although the detection rates look promising, we must note that this database contained only relatively 'easy' face images. More specifically, there was very little occlusion in the images. At present, our system has no mechanism for dealing with feature points that aren't visible. Also, this database only had one subject. Testing with more subjects is desirable in order to evaluate the more general performance of the detector.

## 4.3 Results from Non-marked Database

Since this database is not sequential, we did not use facial tilt correction between frames. As required, many images contain occlusion from spectacles, facial hair and fringe hair. There is also variation in race, age and gender amongst the samples.

In order to evaluate performance, the first 500 images were inspected manually. We counted the images for which 'perfect' detection had been achieved on all points. We also counted 'near-perfect' images, where generally just one feature point was missing or slightly displaced. Examples of these strict criteria are given below:

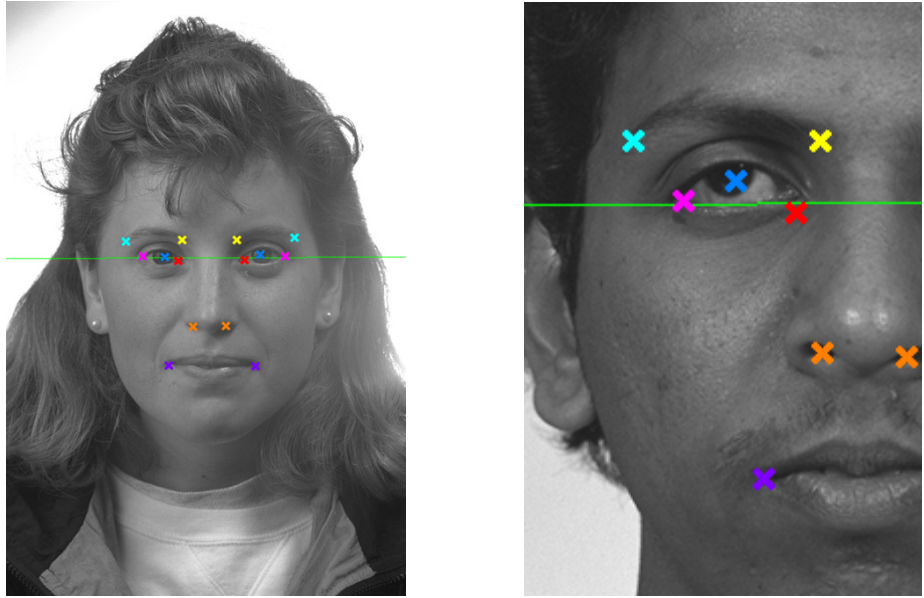

Figure 4.3: *A 'perfect' detection and detail of an imperfect inner eye corner.*

Having all points detected simultaneously is obviously an extremely strict requirement and does not give any specific measure of the performance of each detector. However, we believe our results were encouraging. Perfect detection was achieved 61 times (12.2%) and almost perfect detection a further 86 times (17.2%). Together, these account for almost 30% of cases. We will now look at common ways in which detection failed:

### 4.3.1 Mouths

The huge variability in contortions of the mouth is an ongoing problem. It can be very difficult to define a suitable region of interest. Although not particularly evident using this database is that open mouths were found to regularly fail detection. Training our own mouth detectors will be the subject of future work (see below).

### 4.3.2 Spectacles

Spectacles are hugely problematic. They can lead to feature points, eyes, or even the whole face failing successful detection. Most of the very worst results were due to spectacles. To illustrate this, we present 4 results from the same subject – two with spectacles, then two without.

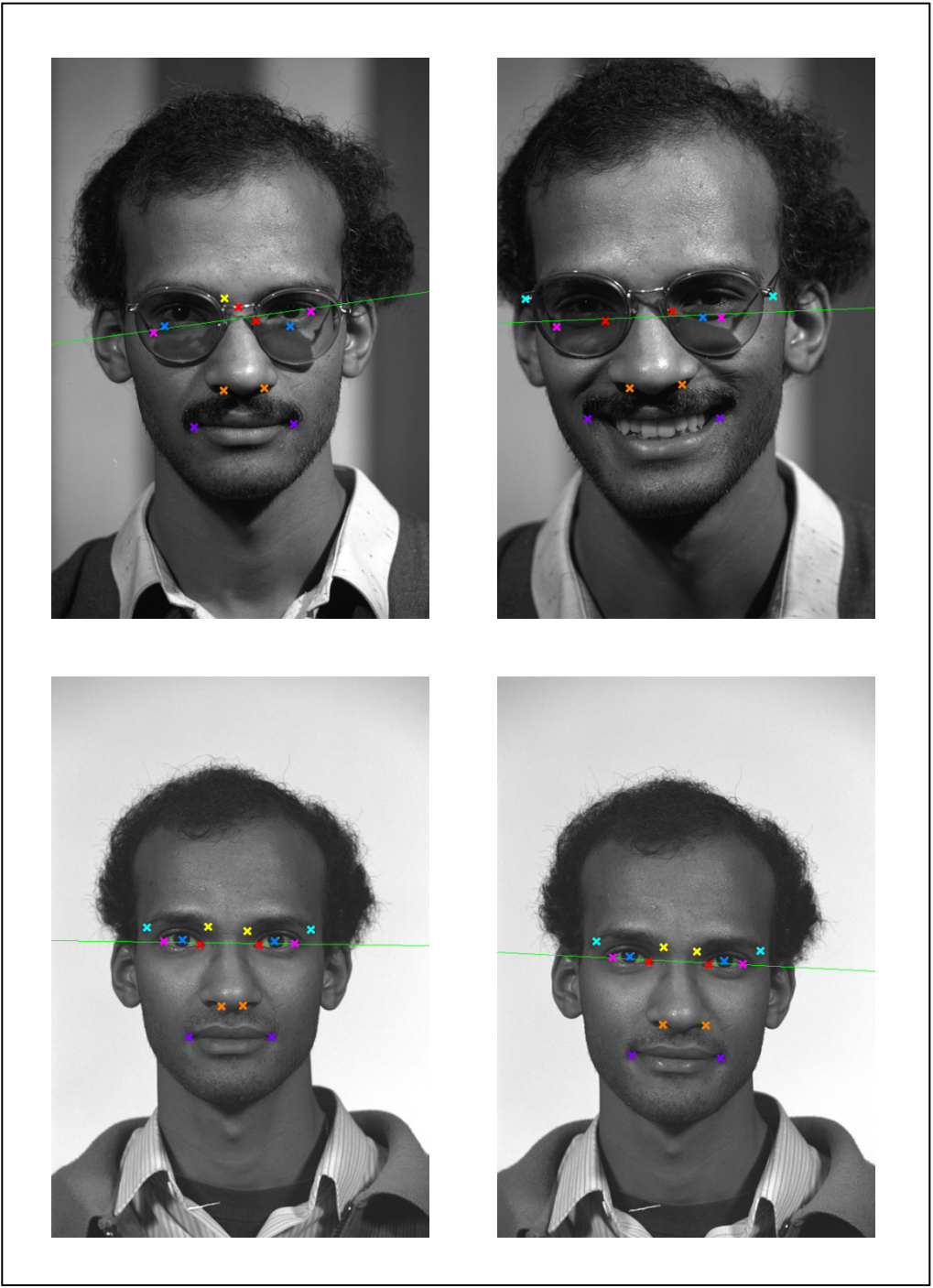

Figure 4.4: *The effect of spectacles on detection.*

### 4.3.3 Fringe Hair

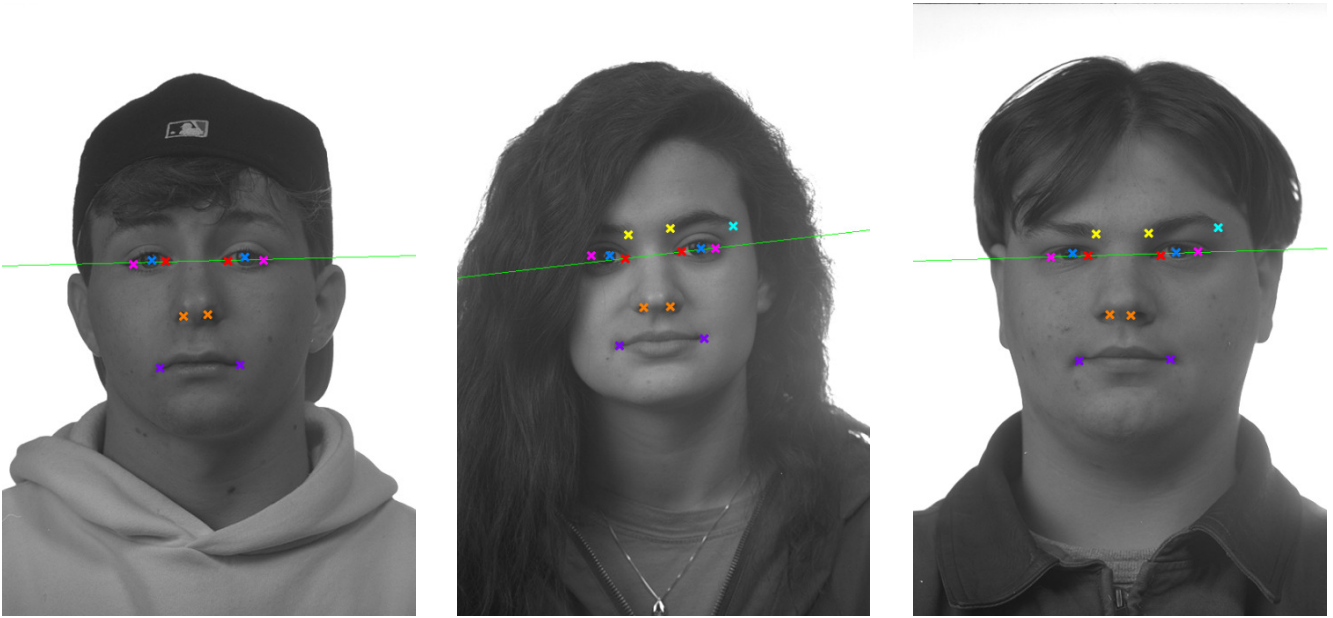

<u>Figure 4.5</u>: *Undetected brow corners in the presence of fringe hair.*

### 4.3.4 Facial Hair

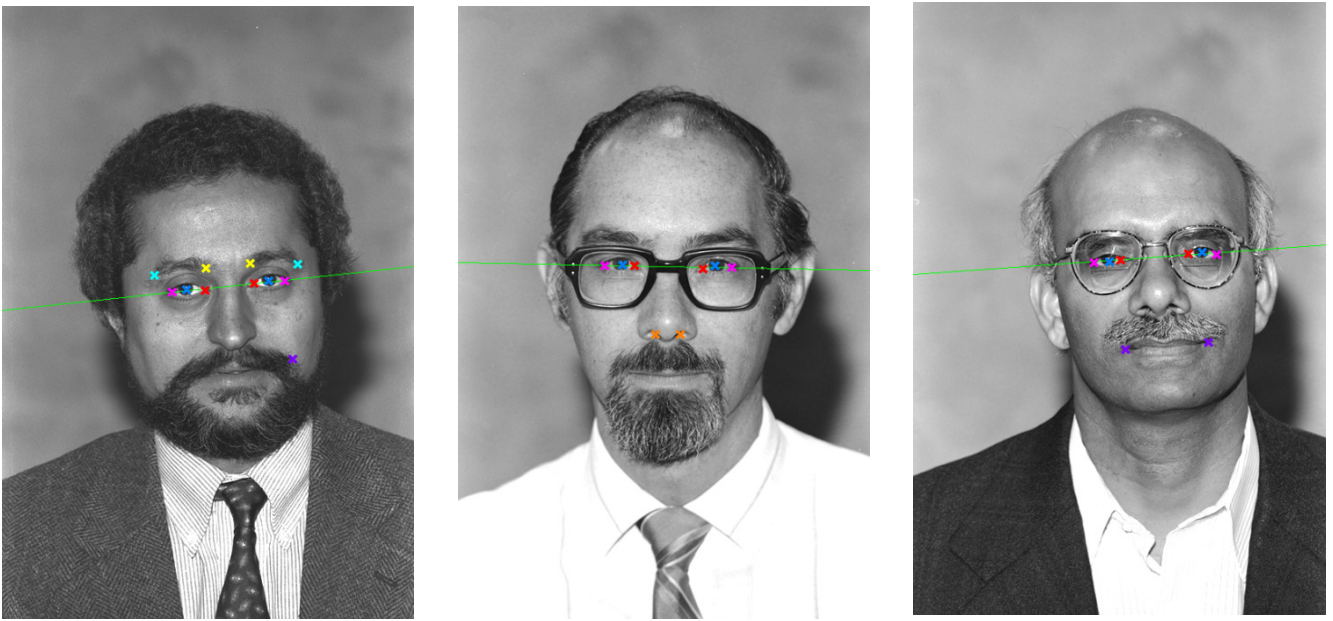

<u>Figure 4.6</u>: *Undetected nostrils and mouth corners in the presence of facial hair.*

# Chapter 5

# Future Work and Conclusions

## 5.1 Future Work

Our system shows good potential and there is great scope for future development. Perhaps the most important would be the incorporation of an explicit geometrical model of the face. In particular, a better understanding of 3D shape would allow our system to deal better with any rotations perpendicular to the image plane. This could perhaps be achieved through a graph matching approach.

Such an approach could help to estimate the location of feature points that are not detected. But, a geometrical model is also particularly relevant for our system because each detector is capable of outputting several candidate coordinates. If some geometric constraints were applied, then it may be possible to select the best overall match from the set of candidate points.

Another approach to this problem would be to manipulate or alter the OpenCV algorithm. Rather than getting multiple outputs that all meet certain criteria, it would be desirable to obtain just one 'best match' in each case. At present, the 'number of neighbours' parameter is used to find better matches, but choosing a specific value prior to detection is limiting – it can lead to the output of zero or several detections in each case. A better result would be to output only the coordinate with most neighbours in each case.

Another important area for future work would be the development of our own feature cascades. The face cascade was deemed to perform well, but feature detectors were somewhat unreliable and did not seem to perform symmetrically. In particular, the mouth detector presents a difficult problem. Analysis of the BioID Database showed that the middle 90% of the width:height ratios of the mouth lay somewhere between 1.4 and 3.5. This is a huge variation, even with that set of natural expressions.

In testing, mouths often failed detection due to facial hair or being open. We suggest that four mouth detectors should be trained to cope with all combinations of beard/no beard and opened/closed. A specific challenge this would present would be obtaining suitable training data. Even then, the great variation in facial hair could cause problems and a new approach may be required.

To address the problems posed by spectacles, it may be beneficial to train a glasses detector. However, since spectacles vary greatly in design and tend to sit differently on different faces, even if we know the person is wearing glasses, this problem may be especially difficult to address. The lenses tend to be non-lambertian (appearance changes depending on angle of view) and the frames are generally opaque, causing occlusion. It may be possible to attempt to fit any available information to a deformable template, thus estimating the occluded parts within the constraints of this template.

We briefly mentioned the possibility of horizontally mirroring detection cascades, rather than having to flip image patches. Upon inspection of the cascade xml files, it indeed seems that this would not be too difficult[1]. Here is an example of how a feature is represented:

```
<feature>
    <rects>
        <_>10 0 2 9 -1.</_>
        <_>7 3 2 3 3.</_></rects>
    <tilted>1</tilted></feature>
    .
    .
    .
```

The rectangles are given in standard OpenCV (x, y, width, height) form. The fifth value is a weighting. So it seems it would just be a case of horizontally flipping these rectangles within whatever detection window we have defined. For example, with our 13x13 window, (10, 0, 2, 9) would become (1,0,2,9).

## 5.2 Conclusions

Our work has identified weaknesses in previous colour-based approaches to feature point detection. Through a careful exploration of human visual perception, we put forward several arguments for an alternative approach, based on learning. We saw how the greater level of abstraction offered by a high-dimensional feature space, combined with correct feature types, could go some way towards mimicking basic neural response and higher-level knowledge.

We then saw how the Viola-Jones method offers a sound basis for building such a detection system. Novel feature sets were seen to offer somewhat more biological responses, but we also noted that some sacrifice may be necessary in order to maintain high frame rate. We thus identified the extended feature set, proposed by Lienhart et al., to be a good compromise.

Our method of training detectors was necessarily meticulous. We sought to normalise tilt and scale in order to optimise the training data. The precise nature of both positive and negative samples was considered in detail. A total of 7 feature point detectors were trained – allowing the detection of 14 points through image patch mirroring. Our code was designed efficiently, to allow new detectors to be incorporated quickly and easily.

Over all, detection results were very encouraging. Testing on an 'easy' sequence of a single talking face, detection rates were largely above 90%. Testing on a much more difficult dataset has also shown promising results and helped us to identify important areas for future work.

---

[1]Thanks to Amanjit Dulai for helping me to understand the cascade xml files.

# Appendix A

## Preparing Training Samples and Implementing a Detector

This document provides a full step-by-step guide to creating our detectors. This method has been specifically tailored to train classifier cascades for the detection of facial feature points. The following C++ projects are required: "Create Logfile of Positives", "Create Database of Negatives" and "RobotVision". The BioID Face Database and ground truths[1] must also be installed and file paths should be updated in the C++ codes.

### 1. Create Positive Training Sample Database Descriptions

This stage requires the C++ project "Create Logfile of Positives". This code reads in the BioID Database ground truths for each image, and then creates a description log of the positive sample(s) related to one of the 20 available feature points. A feature point can either generate one positive sample or samples at several different scales (see Project Report). First we must select the feature and number of sample scales:

| 1.1 | ```enum{ THIS_FEATURE = 7 }; //Right Brow Outer Corner```<br>```enum{ NSCALES = 3 };``` |
|---|---|

<u>Note</u>: We want to describe left/right from our perspective – i.e. the opposite to BioID.

| 1.2 | Executing the code will create the 'log.txt' description file. It should contain 1521 descriptions like this:<br><br>```BioID_0000.pgm 3 184 163 25 25 185 164 23 23 186 165 21 21```<br>```BioID_0001.pgm 3 177 160 23 23 178 161 21 21 179 162 19 19```<br>```.```<br>```.```<br>```.``` |
|---|---|

This format shows the image file name, number of positive samples in that image, and then the (x, y, width, height) description(s) of the rectangular sample(s).

To use with OpenCV, these descriptions have to be converted to a '.vec' file. This is done at the command line, using the OpenCV executable 'createsamples.exe':

| 1.3 | ```createsamples.exe –info log.txt –vec positives.vec –num 4563 –w 13 –h 13``` |
|---|---|

This command takes the descriptions in 'log.txt' and creates 'positives.vec'. In this example, 3 scales were used for each of the 1521 images (giving 4563 total samples). All of these samples are then rescaled to the same dimensions (here, we have 13x13).

<u>Note</u>: *createsamples.exe* can only rescale samples successfully if *iplWarpPerspectiveQ()* is running correctly. This requires the line "```#define HAVE_IPL```" to be added to *_cvhaartraining.h*[2].

## 2. Create Negative Training Sample Image Database

This stage requires the C++ project "Create Database of Negatives". The code generates 16 negative sample images per feature point, randomly located around that feature point (see Project Report).

There are two changes to make in the code:

| | |
|---|---|
| **2.1** | Choose a feature: `enum{ THIS_FEATURE = 15 };` |
| **2.2** | Set the path to save the negative image files: |
| | `string neg_path = "H:\\Left Nostril Training\\bin\\Negatives\\";` |

<u>Note</u>: The number of images generated can be large (typically 22,000). Some systems will hang if you try moving these files around, so it is advisable to set the correct path within the code.

| | |
|---|---|
| **2.3** | Executing the code creates a database of negative sample images (in the path specified) and a log file of image names (in the project folder). |

The log file should be copied to the image directory for the training stage.

## 3. Train Cascade of Classifiers

We now have our positive samples (in a '.vec' file) and our negative samples (a database of images with a text description file). These are now input to the OpenCV Haar-like classifier training function, *haartraining.exe*. This process is computationally intensive and can take days to complete (depending on system performance).

There are a number of input parameters which need to be considered. A typical example is given:

| | |
|---|---|
| **3.1** | `haartraining.exe` **-data** `data/cascade` **-vec** `data/positives.vec` **-bg** `negatives/neg.txt` **-npos** `4563` **-nneg** `24336` **-nstages** `20` **-mem** `2000` **-mode** `ALL` **-w** `20` **-h** `20` **-minhitrate** `0.995` **-maxfalsealarm** `0.5` **-nonsym** |

The most important command-line parameters are:

| | |
|---|---|
| `-data` | Path to save cascade |
| `-vec` | Path to load positive samples '.vec' |
| `-bg` | Path to description of negative samples '.txt' |
| `-npos` | Number of positive samples |
| `-nneg` | Number of negative samples |
| `-nstages` | Number of cascade stages |
| `-nsplits` | Number of tree splits (See [3], pg6) |
| `-nonsym` | Must be set if feature is not left-right symmetrical |
| `-mem` | Physical memory to be used (megabytes) |
| `-minhitrate` | Minimum hit rate |
| `-maxfalsealarm` | Maximum false alarm rate |
| `-mode` | 'BASIC' uses only upright features, 'ALL' uses full set. |
| `-w` and `-h` | Width and height of training samples |

Note: Often the training process will not complete. After a number of stages, it may become impossible for the *minhitrate* and *maxfalsealarm* criteria to be simultaneously achieved.

For example, the command-line output below shows a cascade that has failed to meet training requirements during Stage 6. We can see that increasing the number of features (N) beyond 20 has no further effect on hit rate (HR) and false alarm rate (FA):

```
+----+----+-+---------+---------+---------+---------+
|  N |%SMP|F| ST.THR  |   HR    |   FA    | EXP. ERR|
+----+----+-+---------+---------+---------+---------+
|  18| 79%|-|-1.486423| 0.995035| 0.899966| 0.128105|
+----+----+-+---------+---------+---------+---------+
|  19| 79%|-|-1.434926| 0.995035| 0.872503| 0.155258|
+----+----+-+---------+---------+---------+---------+
|  20| 79%|-|-1.450002| 0.995035| 0.875973| 0.147311|
+----+----+-+---------+---------+---------+---------+
|  21|100%|-|-1.450002| 0.995035| 0.875973| 0.147311|
+----+----+-+---------+---------+---------+---------+
|  22|100%|-|-1.450002| 0.995035| 0.875973| 0.147311|
+----+----+-+---------+---------+---------+---------+
```

When this happens, it is necessary to terminate the program (Ctrl+C) and slacken the training parameters. In general, a small decrease/increase in maxhitrate/minfalsealarm will allow that stage to complete successfully.

It is difficult to say how many stages is 'enough'. This will depend heavily on the required application. In general, extra stages will decrease overall false alarms, but will also decrease hit rate. The best way to evaluate performance is to test the cascade at various stages.

## 4. Convert Cascade

In order to use the cascade for detection, the data first has to be converted to xml format. This is done using *convert_cascade.exe*:

**4.1**  `convert_cascade.exe --size="13x13" data/cascade data/LeftNostril5.xml`

This classifier cascade is now ready to use for detection. OpenCV does have a performance testing tool (*performance.exe*), but this is not relevant for feature point detection due to the constraints we require to be imposed on local background information. Instead, we must implement the detector directly in our C++ code.

## 5. Test Cascade

This stage requires the main C++ feature point detection project "RobotVision". The code uses various classifier cascades to detect feature points within various regions of interest (ROI). Before detection can begin, there are numerous parameters that must first be initialised for every detector. The code has been very carefully designed to make this simple.

Each detector is arranged into a structure whose members need to be initialised. We will use the outer right brow as an example point:

**5.1** The following parameters should be initialised. The values they require are obvious from looking at the code:

| Parameter | Type | Description | Initialise in |
|---|---|---|---|
| m_RightBrowOuter.ROI | CvRect | Rectangular region of interest for detection | main |
| m_RightBrowOuter.cascade | CvHaarClassifierCascade* | Cascade read in from our xml file | constructor |
| m_RightBrowOuter.IsPoint | bool | True for feature points, false for features | constructor |
| m_RightBrowOuter.OnRightSide | bool | True if feature (point) is on the right of the face | constructor |
| m_RightBrowOuter.OK | bool | True if feature (point) has been successfully detected | constructor |
| m_RightBrowOuter.HaarParams | double[4] | Viola-Jones detection parameters* (see below) | constructor |

*Viola-Jones detection parameters:

| Parameter | Definition | Description |
|---|---|---|
| HaarParams[0] | Scale Factor | Haar-like features increase by a scale of `<HaarParams[0]>` on each iteration |
| HaarParams[1] | Min. Neighbours | Positive match requires a cluster with at least `<HaarParams[1]>` neighbours |
| HaarParams[2], HaarParams[3] | Min. Size | Initial Haar-like feature size is `<HaarParams[2]>` x `<HaarParams[3]>` |

With everything correctly initialised, the detection function can now be called from the main function:

**5.2** `DetectFeaturePoint(&m_RightBrowOuter, temp_img);`

Finally, the detected feature point can be displayed by altering the `DrawFeaturePoints` function:

**5.3**
```
if(m_RightBrowOuter.OK)
    DrawCross(m_img, 4, m_RightBrowOuter.point, CV_RGB( 255,255,0)
```

**References (for Appendix A)**: